\documentclass[12pt,twoside]{article}   
\usepackage[super,sort,comma]{natbib}
\usepackage{caption}
\usepackage{subfigure}
\usepackage{amsmath,amssymb,amsfonts}
\usepackage{makecell}

\usepackage{fancyhdr}		

\newcommand{\captionv}[3]{\begin{center}\parbox{#1cm}{\caption[#2]{{\sf #3}}}
        \end{center}}

\usepackage[section]{placeins}   %

\usepackage{graphicx}

\makeatletter \renewcommand\@biblabel[1]{$^{#1}$} \makeatother
\newcommand{\cen}[1]{\begin{center} #1 \end{center}}

\usepackage{hyperref}
\hypersetup{ colorlinks,
    citecolor=blue,
    filecolor=blue,
    linkcolor=blue,
    urlcolor=blue
}

\usepackage{xcolor}

\definecolor{gray}{rgb}{0.6,0.6,0.6}
\definecolor{red}{rgb}{0.85,0,0}
\definecolor{green}{rgb}{0,0.85,0}
\definecolor{blue}{rgb}{0,0,0.85}
\definecolor{beige}{rgb}{0.92,0.87,0.78}
\usepackage[all]{hypcap}    

\begin{document}

\cen{\sf {\Large {\bfseries A Structure Feature Extraction Algorithm for Multi-modal Forearm Registration } \\  
\vspace*{10mm}
$^1$Jiaxin Li, $^1$Yan Ding, $^1$Weizhong Zhang, $^2$Yifan Zhao, $^3$Lingxi Guo, $^3$Zhe Yang} \\

$^1$Key Laboratory of Dynamics and Control of Flight Vehicle, Ministry of Education,
School of Aerospace Engineering, Beijing Institute of Technology, Beijing 100081,
China

$^2$ School of Aerospace, Transport and Manufacturing, Cranfield University, Wharley End, Bedford, UK

$^3$ Science and Technology on Space Physics Laboratory ,Beijing 100076,PR China
\vspace{5mm}\\
Version typeset \today\\
}

\pagenumbering{roman}
\setcounter{page}{1}
\pagestyle{plain}
Corresponding Author: Yan Ding. email: dingyan@bit.edu.cn\\

\begin{abstract}
\noindent {\bf Purpose:} Augmented reality technology based on image registration is becoming increasingly popular for the convenience of pre-surgery preparation and medical education. This paper focuses on the registration of forearm images and digital anatomical models. Due to the difference in texture features of forearm multi-modal images, this paper proposes a forearm feature representation curve (FFRC) based on structure features to extract feature points of forearm images. In addition, we create an FFRC-compliant multi-modal image registration framework (FAM) for the forearm.\\ 
{\bf Methods:} FFRC counts the number of feature pixels in each column of the forearm binary image and uses the Kalman filter to eliminate data deviation. FAM screens matching spots using the FFRC feature of forearm pictures before applying affine transformation. FAM-TPS, which adds Thin Plate Spline to FAM, performs deformable registration by using the matched points as the interpolation function's control points. FAM and FAM-TPS are evaluated on a dataset containing the axial rotation of the forearm. We employ Dice coefficient (DC), Jaccard coefficient (JC), Hausdorff distance (HD), Average surface distance metric (ASD), Average symmetric surface distance (ASSD) indicators to evaluate the registration accuracy of the framework and use Fréchet Inception Distance (FID) and Euler distance (ED) to evaluate the accuracy of the feature point matching.\\
{\bf Results:}Other classic feature-based registration methods were compared to FAM and FAM-TPS. In the registration experiment, our framework (FAM) achieved the DC of 0.987, JC of 0.974, HD of 387.21, ASD of 1.22, ASSD of 4.02, by adding thin-plate spline interpolation, our framework (FAM-TPS) achieved the DC of 0.991, JC of 0.982, HD of 395.69, ASD of 0.958, ASSD of 3.11. Furthermore, tests of resilience showed that our framework can register forearm images with varying rotation angles.\\
{\bf Conclusions:} The multi-modal forearm feature points can be retrieved properly using FFRC, and the forearm's multi-modal image registration can be completed using the affine transformation matrix and thin-plate spline interpolation. Furthermore, the rotation angle of the forearm has no effect on the registration effects of FAM and FAM-TPS, and the peak value of the feature curve based on the structure has a correlation with the rotation angle of the forearm. \\

\end{abstract}

\newpage     

\tableofcontents

\newpage

\setlength{\baselineskip}{0.7cm}      

\pagenumbering{arabic}
\setcounter{page}{1}
\pagestyle{fancy}
\section{Introduction}
Soft tissue injuries are widespread in sports, and they are particularly common in soccer\cite{ekstrand2011epidemiology}, rugby\cite{lopez2012profile}, basketball\cite{borowski2008epidemiology}, track, and field\cite{jacobsson2012prevalence}. The mechanism of injury might be direct, indirect, or mixed trauma\cite{lopez2012profile,borowski2008epidemiology} and can result in a disability that will take surgeries to repair. Preoperative planning is required for some critical conditions, such as tendon rupture\cite{burnham2011technique}. Using the forearm as an example, the augmented reality (AR) technology based on image registration can project the digital anatomical model on the forearm image, which is convenient for the location of the injury location and the formulation of the surgical plan. Image registration lies in the core of AR, which aligns the virtual scene with reality.  As a result, image registration accuracy is crucial for surgical planning, which leads to a variety of image registration approaches.
\par
Image registration is the process of transforming different image datasets into one coordinate system with matched imaging contents, which has significant applications in the medical image processing field. However, for non-rigid objects registration, deformable multi-modal registration methods are needed to be introduced to eliminate the gap that cannot be avoided by rigid registration algorithms. A deformable registration strategy usually consists of two sequential steps: one is a globally aligned affine transformation, and then a deformable transformation. We concentrate on the first step, in which we design a feature point extraction and matching method based on the forearm structure.
\par
Affine transformation needs to calculate the affine transformation matrix between the two paired images, which's meant to find at least four pairs of matching points between the two images. An image matching framework usually consists of three major parts: feature detection, feature description, and matching methods. For image matching, finding an appropriate feature descriptor is the most important and challenging step. Descriptors based on various features have emerged recently. Gradient statistic approaches are often used to form float type descriptors such as the histogram of oriented gradients (HOG)\cite{dalal2005histograms} as presented in SIFT\cite{lowe1999object,lowe2004distinctive}. In SIFT, feature scale and orientation are respectively determined by DoG computation and the largest bin in a histogram of gradient orientation from a local circular region around the detected keypoint, thus achieving scale and rotation invariance. Another representative descriptor, namely, SURF\cite{bay2006surf}, can accelerate the SIFT operator by using the responses of Haar wavelets to approximate gradient computation than SIFT. However, these traditional algorithms are often difficult to work in the situation where the source image and the target image are quite different. This requirement has prompted investigations on learning-based descriptors, which have recently become dominantly popular due to their data-driven property and promising performances. In general, existing methods based on learning consists of two forms, namely, metric learning\cite{weinberger2009distance,zagoruyko2015learning,han2015matchnet,kedem2012non,wang2017deep} and descriptor learning\cite{salti2015learning,balntas2016pn,zhang2017learning,mishchuk2017working,wei2018kernelized,he2018local,tian2019sosnet,luo2019contextdesc}, according to the output of deep learning-based descriptors. 
\par
Despite the existence of quite accurate feature point extraction algorithms, they still suffer from a lack of robustness, which is a critical aspect for AR. As a result, the robustness of multi-modal forearm registration is the emphasis of this work. The inconsistency of the textual expression of the multi-modal forearm image and the registration stability of the forearm axial rotation are the two key issues. Hence, we proposed a forearm feature representation curve (FFRC) based on the structural feature of the forearm. This curve is not only unaffected by the textural feature, but it can also represent the axial rotation angle of the forearm. Moreover, we designed a forearm registration framework (FAM and FAM-TPS) based on FFRC and used a variety of indicators to verify the stability of our framework.






\section{Methods}
\begin{figure}[htbp]
   \begin{center}
   \includegraphics[width=15cm]{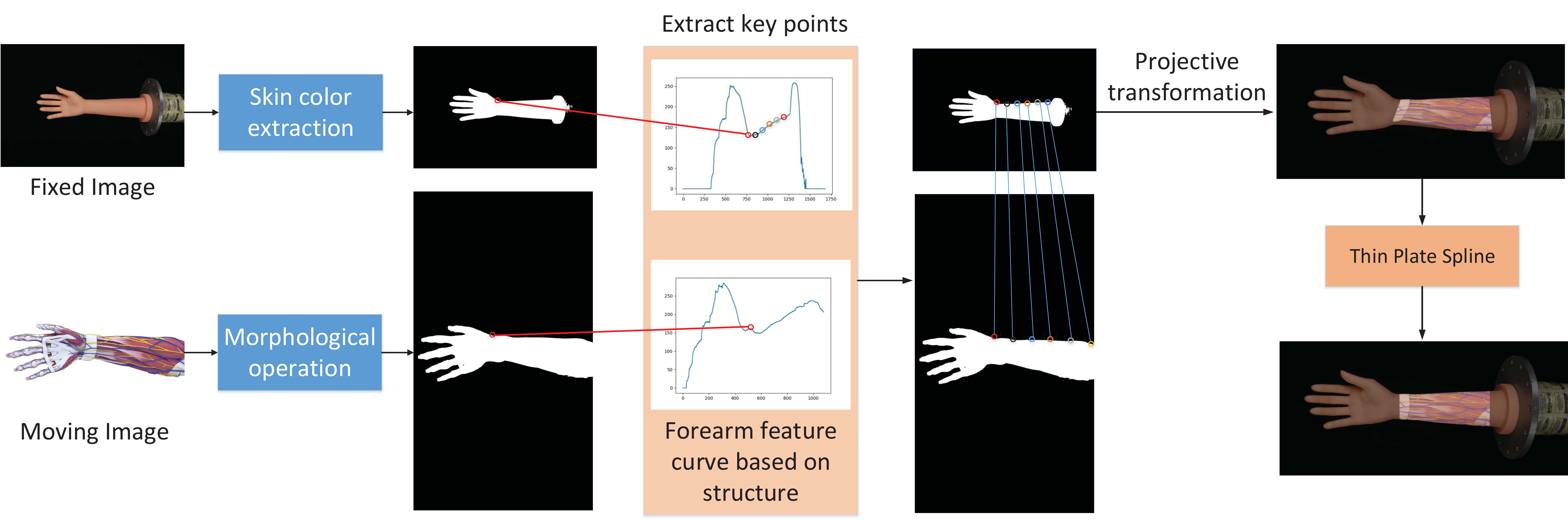}
   \captionv{12}{Short title - can be blank}
   {\textbf{The registration framework based on FFRC.} This figure depicts a multi-modal registration framework based on FFRC. From left to right, extract the mask of the forearm, the forearm representation curve, the feature point matching, and the deformable registration.\label{figure_1} 
    }  
    \end{center}
\end{figure}

The forearm image is defined as a fixed image in our registration framework, as shown in Figure \ref{figure_1}, while the digital anatomical model is defined as a floating image. The skin color extraction method and the morphological method are then used to extract their binary maps. After rotating the paired binary maps to the level and computing the main direction, we propose the forearm feature representation curve for extracting their matching feature points. Finally, the feature points are used for deformable registration including affine transformation and thin-plate Spline methods.
\subsection{Skin Color Feature Extraction}
Skin color feature extraction is to extract the mask of the forearm region from the image, which can provide a basis for extracting feature points of the forearm. The forearm region in the image is segmented using the YCrCb color space and OTSU threshold segmentation.
\par
The YCrCb color space is a typical skin color detection color model, where Y represents brightness, Cr represents the red component in the light source, and Cb represents the blue component in the light source. The display range of human skin color in the YCrCb space is confined in a limited portion of the Cr channel. As a result, we extract the Cr component from the image and perform threshold segmentation through OTSU, resulting in a well-segmented human skin color area. The process of skin color extraction can be represented as:
\begin{equation}
S(\mathbf{F_o})=OTSU(G(\mathbf{F_o(Cr)})),
\end{equation}
where $\mathbf{F_o}$ denotes the original image of the forearm, $S(\cdot)$ denotes the skin color feature extraction result, $\mathbf{F_o(Cr)}$ denotes the Cr component of the forearm image, $G(\cdot)$ denotes a 5x5 Gaussian filter, $OTSU(\cdot)$ denotes the adaptive threshold processing method based on maximum between-class variance.
\par
The extraction result is shown in Figure \ref{figure_3}. By comparing the forearm image and its extracted mask, we can observe that the algorithm is able to extract the forearm from the background.
\subsection{Binary Map Principal Direction Correction}
\begin{figure}[htbp]
   \begin{center}
   \includegraphics[width=10cm]{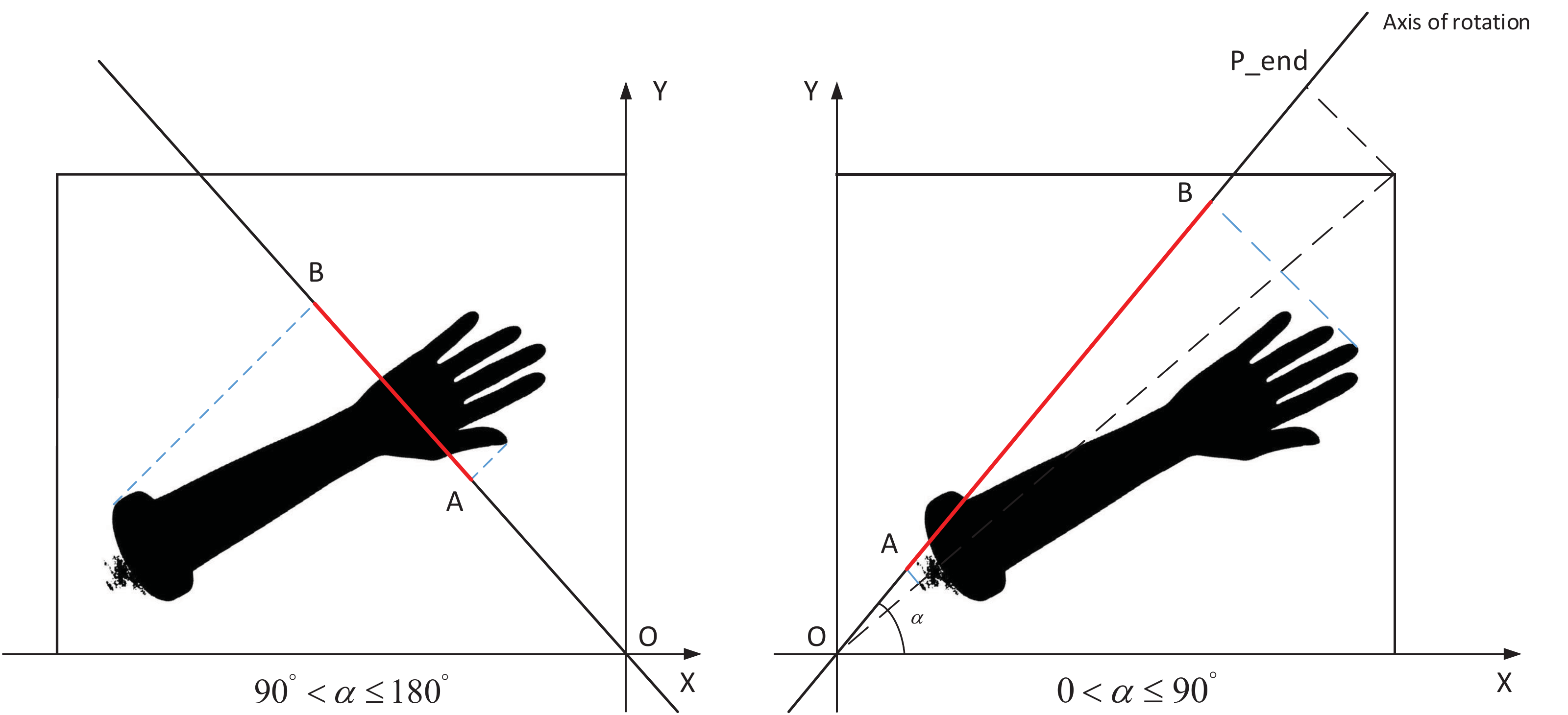}
   \captionv{12}{Short title - can be blank}
   {\textbf{Extraction of main direction for the forearm.} This picture shows the extraction process of the main direction of the binary image (For visual display, the forearm is turned to black.). The picture on the right shows the angle of the rotation axis between 0-90°, and the picture on the left shows the angle of the rotation axis between 90-180°. The AB line segment represents the length of the binary image projected onto the axis of rotation.\label{figure_2}
    }  
    \end{center}
\end{figure}
The main direction of the binary map notes that the angle between the trend direction of the pixel point distribution in the image and the X-axis. The binary image can be parallel to the X-axis according to the main direction angle after determining the main direction of the binary map including the forearm. This operation can accelerate the extraction speed of forearm feature points in the third section.
\par
In this paper, the main direction is limited in the range of $[0,180^{\circ})$. As shown in Figure \ref{figure_2}, we rotate a straight line once within this range, and respectively calculate the length of the projection line segment $AB$(\ref{equ:AB}) where the point of interest in the image falls on this axis of rotation.
\begin{equation}
\begin{array}{l}
A={Min}\left(\left|P_{r}\right|\right) \\
B={Max}\left(\left|P_{r}\right|\right)
\end{array}
\label{equ:AB}
\end{equation}
where $P_r$ denotes the projection point of the binary graph on the rotation axis, $|\cdot|$ denotes the distance from the projection point to the origin.
\par
Since the rotation axis passes through the origin, the linear equation can be represented by $(y=kx)$(except for $90^{\circ}$), and the projection coordinates of the point on the image to the rotation axis can be easily calculated. The projection line segment falling on the rotation axis has a maximum value during the rotation process when the rotation axis is at an angle, and this angle is the major direction. When calculating the length of the projection line segment, if the rotation angle of the rotation axis is greater than 90 degrees, we translate the image along the X-axis to ensure that the projection line segment is always above the Y-axis.
\par
Simultaneously, in order to speed up this phase, we use a method that is easier to calculate but loses some precision. Since the forearm can be approximately regarded as a rigid body, after obtaining the binary map of the forearm, we extract the minimum circumscribed rectangle of the forearm and directly treat the rotation angle of the minimum circumscribed rectangle as the main direction of the binary map.
\subsection{Forearm Feature Representation Curve}
\begin{figure}[htbp]
   \begin{center}
   \includegraphics[width=15cm]{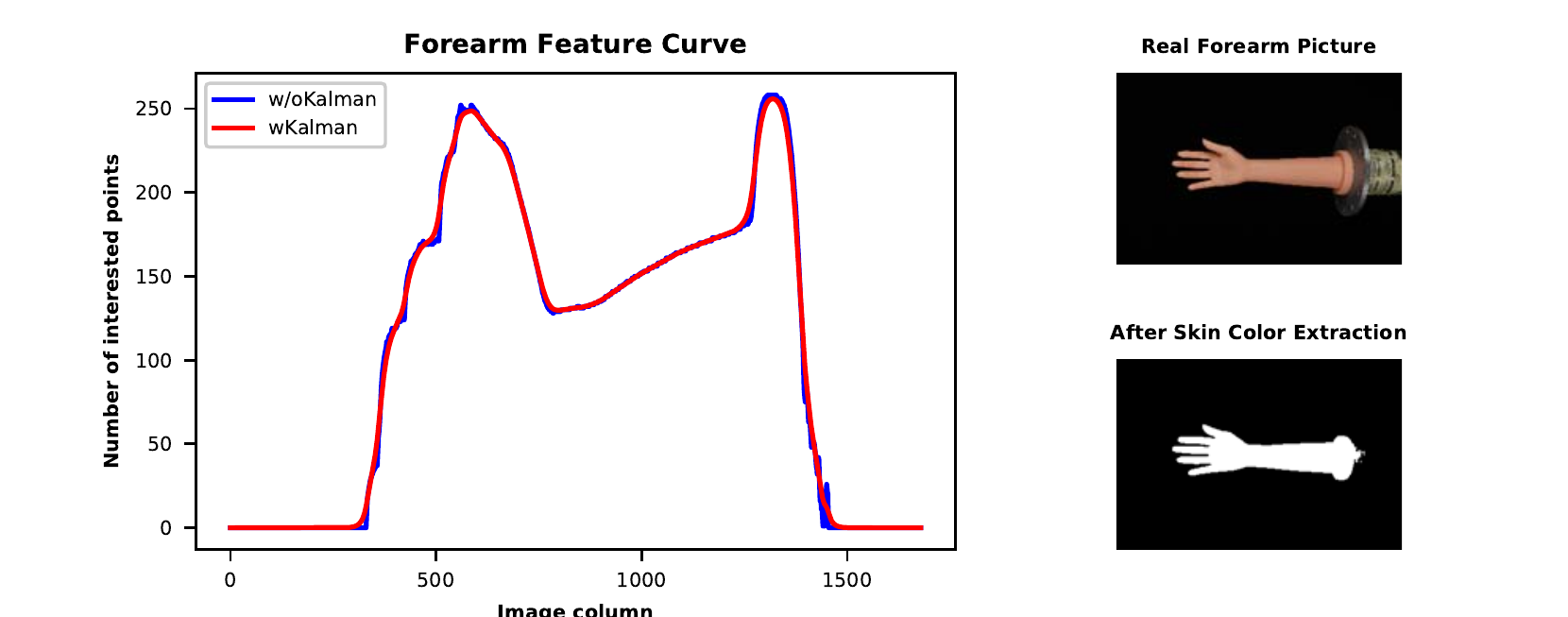}
   \captionv{12}{Short title - can be blank}
   {\textbf{Forearm Feature Representation Curve.} The figure on the left shows the forearm feature curve extracted by the structure-based feature extraction method. The blue curve represents the original curve, and the red curve represents the Kalman filtered curve. The image on the right shows the result of skin color extraction.\label{figure_3} 
    }  
    \end{center}
\end{figure}
After skin color extraction and the main direction extraction of the binary image, we can get the forearm mask parallel to the X-axis of the image. We count the total number of points in each column of the image with a pixel value of 255 and construct a one-dimensional curve $C_o$, as represented by the blue curve in Figure \ref{figure_3}. $C_o$ can be expressed as:
\begin{equation}
\begin{array}{l}
C_{o}(i)=\sum_{j=0}^{h}\left(S^{\prime}\left(\mathbf{F_{o}}\right)_{x=i}^{y=j}\right), 0 \leq i \leq w \\
\end{array}
\end{equation}
where $\mathbf{F_o}$ denotes the original image of forearm, $S^{\prime}\left(\mathbf{F_{o}}\right)$ is expressed in formula (\ref{equ:S(F)_prime}), $w$ denotes the width of $\mathbf{F_o}$, $h$ denotes the height of $\mathbf{F_o}$. 
\begin{equation}
S^{\prime}\left(F_{o}\right)=\left\{\begin{array}{ll}
1, & \text{if } S\left(\mathbf{F_{o}}\right)>0 \\
0, & \text{if } S\left(\mathbf{F_{o}}\right)=0
\end{array}\right.
\label{equ:S(F)_prime}
\end{equation}
It can be seen that the location of the wrist of the forearm corresponds to the position of the valley in this feature curve $C_o$. The blue curve is unsmooth and there is noise interference due to the presence of burrs on the edge of the forearm mask. Therefore, Kalman filter is used to process the original features $C_o$ to obtain the filtered features $C_k$.
\par
\begin{figure}[htbp]
   \begin{center}
   \includegraphics[width=10cm]{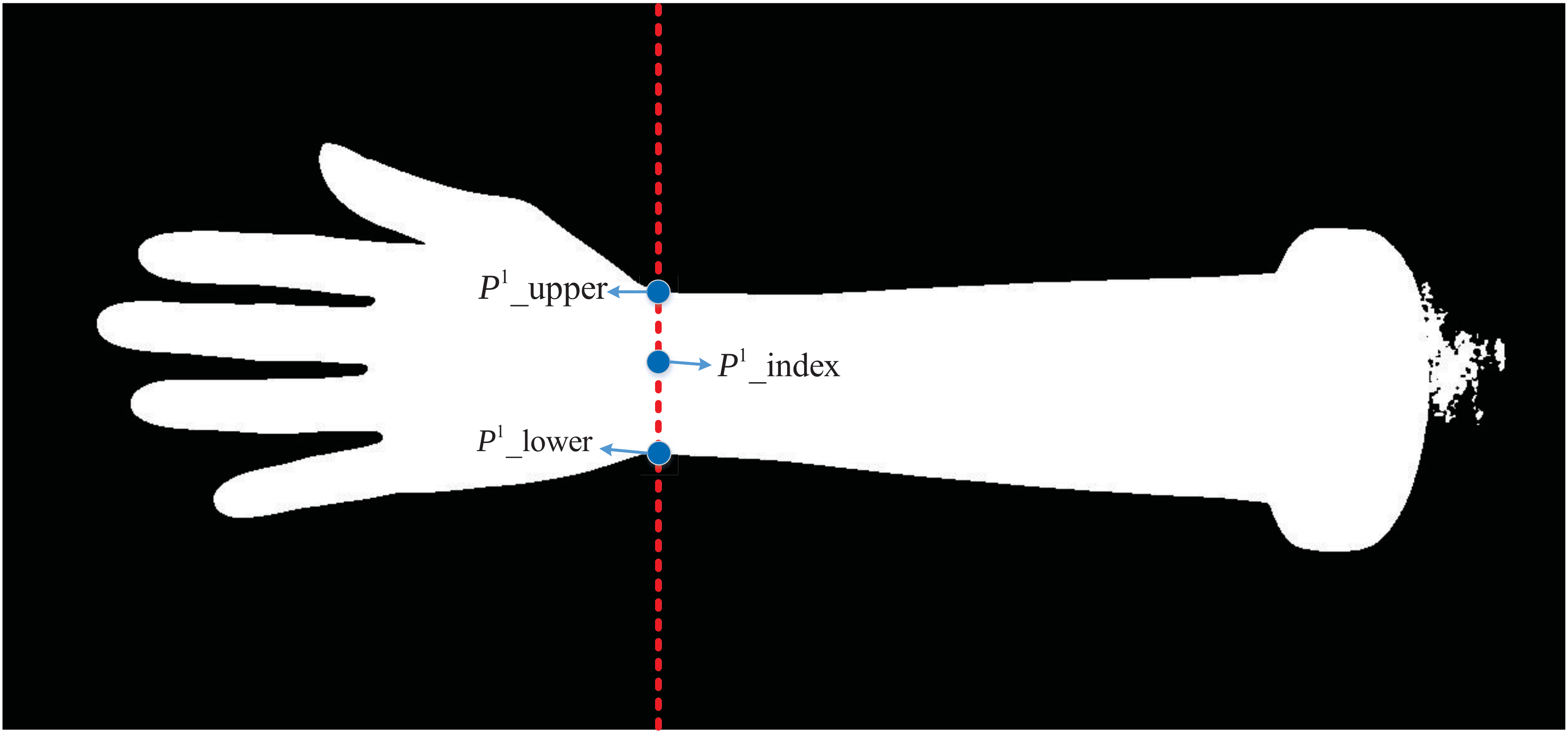}
   \captionv{12}{Short title - can be blank}
   {\textbf{Feature point search scheme.} This picture shows how to find the upper and lower edges of the forearm through the index of the feature point. The red dotted line represents the search path.
   \label{figure_4} 
    }  
    \end{center}
\end{figure}

The trough of the curve can be filtered and estimated using Eq.(\ref{equ:P_w}), which yields the trough corresponding to the wrist X-axis location ${P_x}^1$ based on the forearm feature representation curve.
\begin{equation}
P_{x}^w=\text{Argmax}\left(\sum_{i=1}^{L} \text{sign}\left(C\left(t_{j}-\frac{L}{2}+i\right)-C\left(t_{j}\right)\right)\right) \mid j=1,2 \ldots, n
\label{equ:P_w}
\end{equation}
where $P_{x}^w$ denotes the x coordinate of the key point of the wrist, $sign(\cdot)$ denotes expressed by the formula\ref{equ:sign}. $C(\cdot)$ denotes the feature value in $C_k$, $t_j$ denotes the x coordinate of the $j-th$ valley point, L denotes a hyperparameter related to image width.
\begin{equation}
\operatorname{sign}(x)=\left\{\begin{array}{ll}
1, & x>0 \\
0, & x \leq 0
\end{array}\right.
\label{equ:sign}
\end{equation}
Concurrently, the most distal point of the forearm mask from the wrist point is considered as another feature point ${P_x}^n$. We uniformly interpolate and sample the two locations based on the structural feature of the forearm, but we only have the column coordinates of these spots at this time. With the convenience of the forearm mask being parallel to the X-axis of the image, we can directly obtain the point coordinates of the upper and lower edges through the change of pixel value just as Figure \ref{figure_4}. 
These extracted forearm boundary points are regarded as feature points of the forearm image. It should be noted that the actual forearm length corresponding to different forearm imaging models is the same for one person, so these feature points are also in an one-to-one correspondence in practice. This completes the extraction and matching of feature points.
\subsection{Deformable Registration Framework}
\begin{figure}[htbp]
   \begin{center}
   \includegraphics[width=12cm]{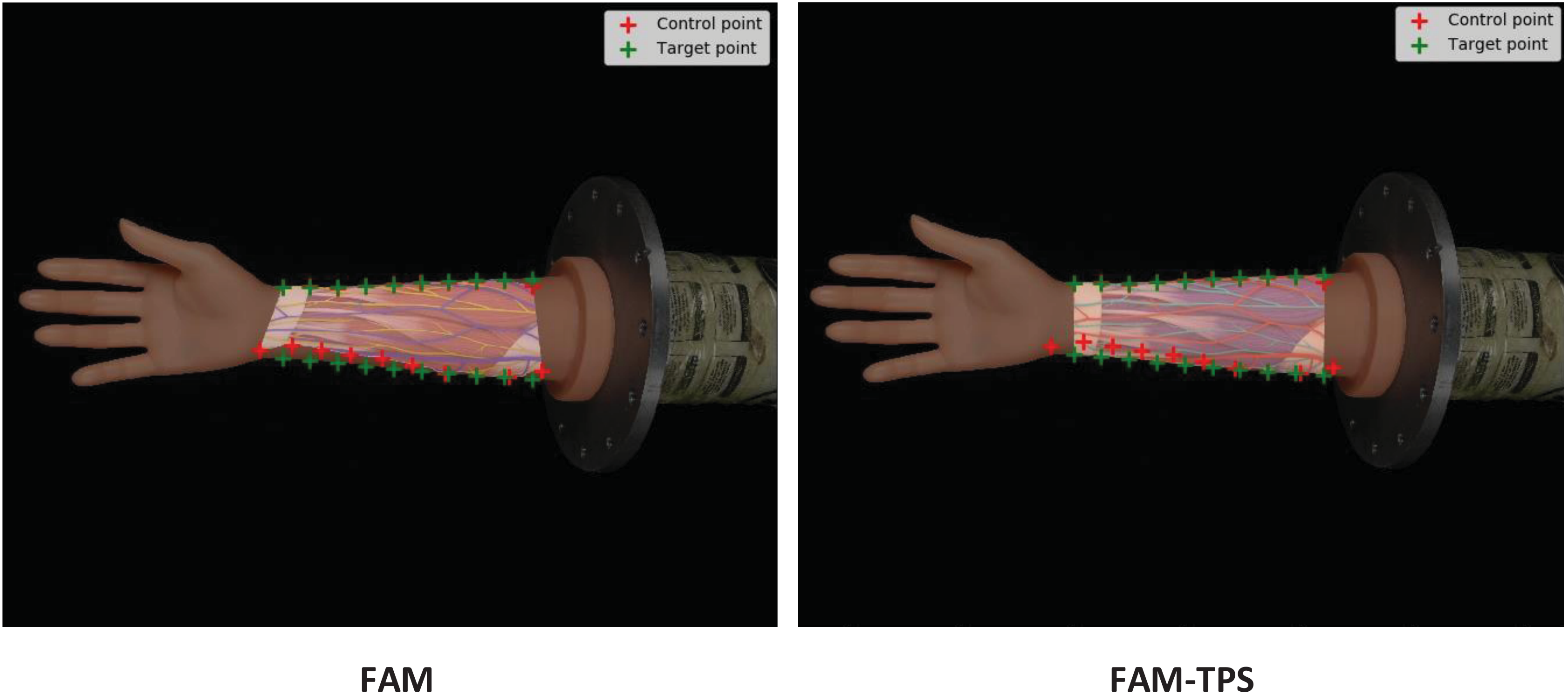}
   \captionv{12}{Short title - can be blank}
   {\textbf{Non-rigid registration.} This picture shows the method of selecting control points for thin-plate spline interpolation. The red cross indicates the control point, and the green cross indicates the target position. 
   \label{figure_5} 
    }  
    \end{center}
\end{figure}
\begin{figure}[htbp]
   \begin{center}
   \includegraphics[width=13cm]{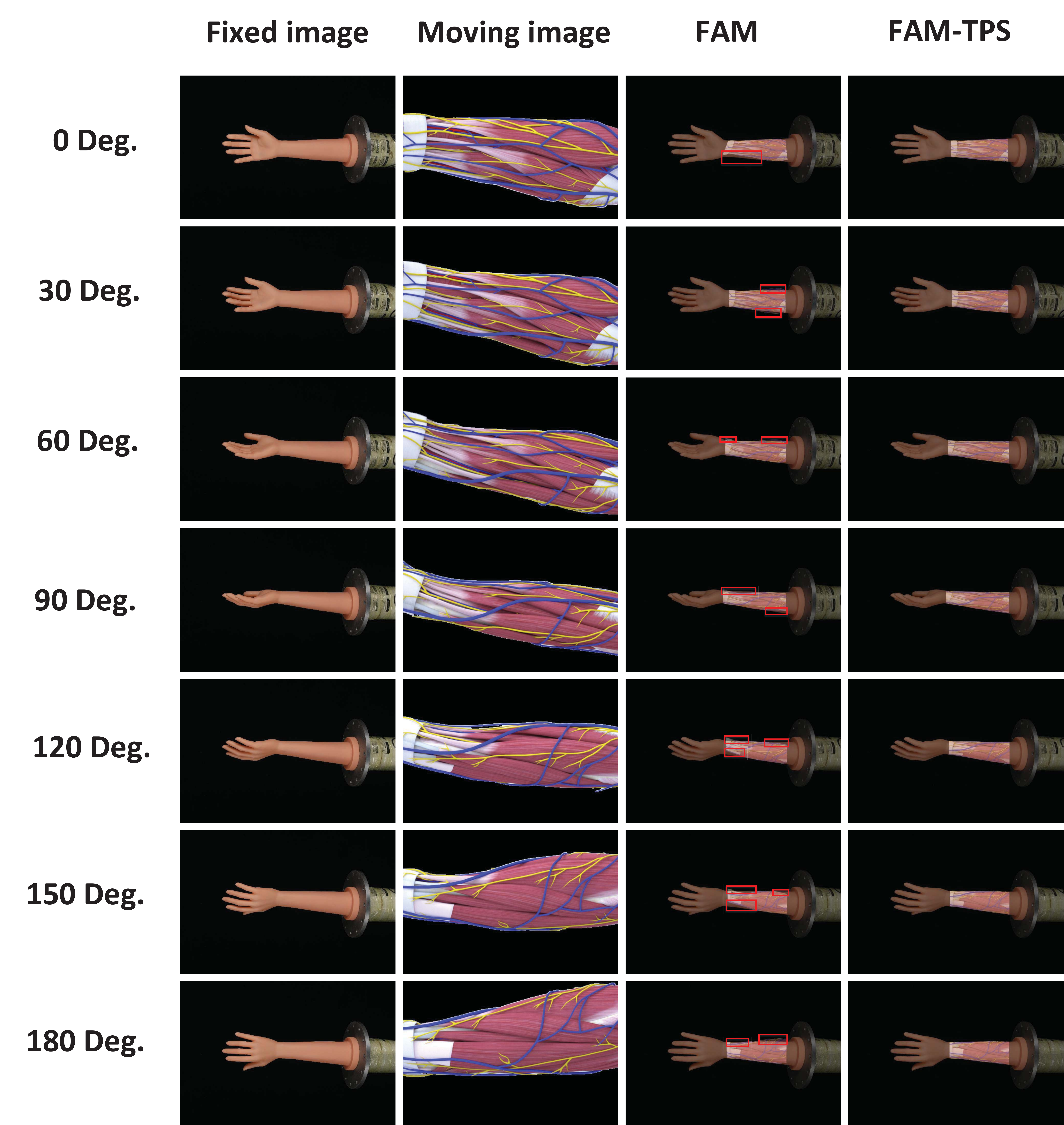}
   \captionv{12}{Short title - can be blank}
   {\textbf{Results of forearm registration with different rotation angles.} From left to right, the fixed image, moving image, FAM rigid registration image, and FAM-TPS deformable registration image are shown in the forearm registration process. From top to bottom, it represents the image group where the forearm axis rotates 0-180°, sampling every 30 degrees. The red rectangle in the third column represents the edge defect of rigid registration. The registered image is a weighted superposition of the fixed image and the transformed floating image (the weight of fixed image is 0.4, the weight of moving image is 0.6). 
   \label{figure_6} 
    }  
    \end{center}
\end{figure}

When performing deformable registration, the matrix $H$ of the affine transformation of the two images is calculated by the matching points extracted by FFRC. Hence, floating images can be mapped to fixed images by the affine transformation. The result of rigid registration is shown in Figure \ref{figure_6}. Subsequently, we use these matching points as the control points of the thin plate spline interpolation for deformable registration and finally, obtain the deformed registration result. The selection of control points is shown in Figure \ref{figure_5}. According to the physiological structure of the forearm, the green points are regarded as control points, and the corresponding red points are regarded as target positions. It can be observed that the anatomical image of the forearm fits well with the forearm image. The specific evaluation criteria will be explained in the experimental chapter.

\newpage     

\section{Experiments}
Several experiments are conducted to demonstrate the capabilities of FAM in terms of its robustness and general applicability to medical forearm image registration. We have implemented several classic descriptor extraction methods and registered forearm images under the same framework. First, we use DC (Dice coefficient), JC (Jaccard coefficient), HD (Hausdorff Distance), ASD (Average surface distance metric), and ASSD (Average symmetric surface distance) to evaluate the registration effect of different methods. Second, by using the forearm image group that rotates 360 degrees axially, the Euler distance between the projection of the feature point and the ground-truth was tested to the accuracy of matching points, and the FID of the image before and after registration was tested to verify the accuracy of registration results. Finally, when the forearm was rotated in the axial direction, experimental results validated that the feature peak value and the rotation angle of the forearm have an obvious corresponding relationship.
\subsection{Registration methods comparison experiment}
To demonstrate the superiority of our proposed algorithm, all registration methods compared with FAM were tested on the forearm registration dataset in the same registration framework. The registration methods based on image descriptors such as SIFT, BRISK\cite{leutenegger2011brisk}, SURF, AKAZE\cite{alcantarilla2011fast}, ORB\cite{rublee2011orb} are used for comparison. Using the DC, JC, HD, ASD, and ASSD indicators to evaluate the registered images, the final displayed results are shown in Table \ref{tab_methods_eval}.
\par
The registration framework first extracts the feature points of the fixed image and the moving image through different methods and match them with the Flann matcher. Then, it calculates the affine transformation matrix between the two images and completes the image registration through affine transformation. In terms of parameters setting, SIFT and SURF algorithms use KD tree for nearest neighbor matching, ORB, BRISK and AKAZE algorithms use LSH for nearest neighbor matching, and both recursive 50 times. When using the extracted matching point pairs to calculate the affine transformation matrix of the two images, the RANSAC method is used to eliminate the wrong matching points, and the threshold parameter is set to 4. 
\begin{table}[htbp]
\begin{center}
\captionv{10}{}{Overall registration result for the evaluated methods. 
Seven descriptor extraction methods were tested, including the two methods we proposed. Five evaluation criteria are used to evaluate the accuracy of the registration results of these methods. For DC and JC indicators, larger is better. For HD, ASD, and ASSD indicators, smaller is better. Bold font indicates the best result.
\label{tab_methods_eval}
\vspace*{2ex}
}
\begin{tabular}{llllll}
\hline\noalign{\smallskip}
\multicolumn{1}{l}{} &\multicolumn{5}{l}{Evaluation Standards} \\
\hline\noalign{\smallskip}
Methods & $DC\uparrow$ & $JC\uparrow$ & $HD\downarrow$ & $ASD\downarrow$ & $ASSD\downarrow$\\
\hline\noalign{\smallskip}
SIFT & \makecell[r]{0.47} & \makecell[r]{0.37} & \makecell[r]{485.21} & \makecell[r]{146.93} & \makecell[r]{155.11}\\
BRISK & \makecell[r]{0.31} & \makecell[r]{0.18} & \makecell[r]{515.37} & \makecell[r]{160.60} & \makecell[r]{188.74}\\
SURF & \makecell[r]{0.50} & \makecell[r]{0.38} & \makecell[r]{447.64} & \makecell[r]{196.74} & \makecell[r]{123.58}\\
AKAZE & \makecell[r]{0.56} & \makecell[r]{0.45} & \makecell[r]{447.38} & \makecell[r]{71.93} & \makecell[r]{90.83}\\
ORB & \makecell[r]{0.28} & \makecell[r]{0.16} & \makecell[r]{516.71} & \makecell[r]{230.29} & \makecell[r]{235.96}\\
FAM(Ours) & \makecell[r]{0.987} & \makecell[r]{0.974} & \makecell[r]{\textbf{387.21}} & \makecell[r]{1.22} & \makecell[r]{4.02}\\
FAM-tps(Ours) & \makecell[r]{\textbf{0.991}} & \makecell[r]{\textbf{0.982}} & \makecell[r]{395.69} & \makecell[r]{\textbf{0.958}} & \makecell[r]{\textbf{3.11}}\\
\hline
\end{tabular}
\end{center}
\end{table}
\par
It can be observed from the table that our method is the best among other registration algorithms on descriptors in all indicators. At the same time, based on the matching points extracted by our FAM method, the registration accuracy of the registered image after the thin-plate spline interpolation can be further improved.
\subsection{Axial rotation registration}
\begin{figure}[t]
   \begin{center}
   \includegraphics[width=10cm]{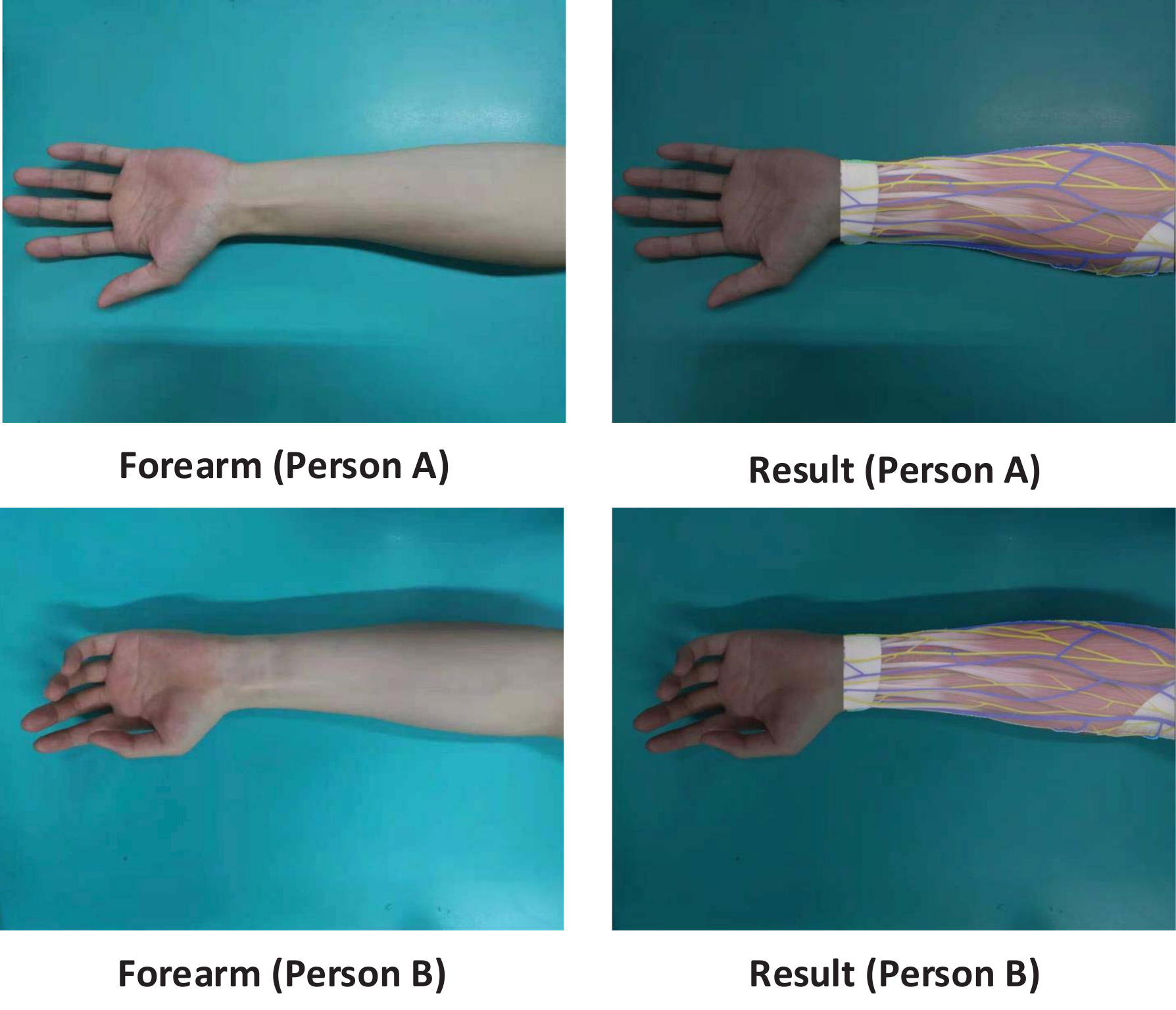}
   \captionv{12}{Short title - can be blank}
   {\textbf{Real forearm registration result.} The left part is the forearm images of different people, and the right part is the result of the registration of the FAM-TPS algorithm we proposed.
   \label{figure_11} 
    }  
    \end{center}
\end{figure}

The forearm axial rotation dataset contains fixed images and moving images in which the forearm rotates 360 degrees around the axis, a set of image pairs every 5 degrees, a total of 72 image pairs. We conducted a registration experiment on this dataset, and tested the registration results of FAM and FAM-TPS, as shown in Figure \ref{figure_6}. It can be seen that the affine transformation matrix calculated by FAM can  register the fixed image with the moving image, but there exist gaps at the edge of forearms. To solve this problem, we use the two image feature point pairs extracted by FAM as control points to perform thin-plate spline interpolation. It can be seen that the registration effect of FAM-TPS is better than FAM, and the gaps at the edges of the image are also eliminated. In addition, FAM-TPS has also been tested for registration on the forearms of different people. As shown in Figure \ref{figure_11}, our algorithm can still register successfully even if the shape of the forearms is quite different.
\begin{figure}[htbp]
   \begin{center}
   \includegraphics[width=10cm]{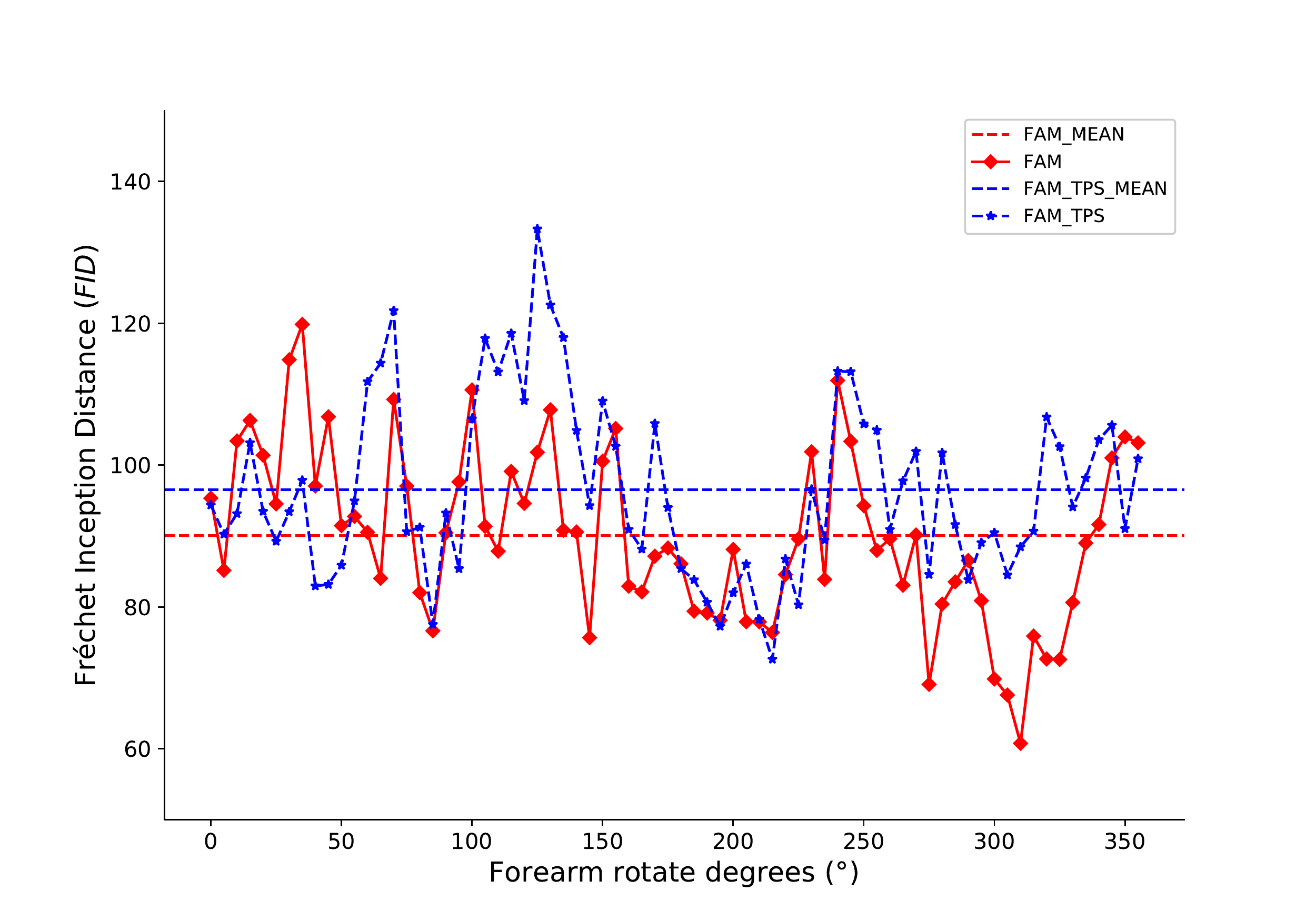}
   \captionv{12}{Short title - can be blank}
   {\textbf{The FID registration evaluation result of the forearm.} The figure shows the registration accuracy curve of the forearm one revolution in the axial direction. The red solid line represents the FID curve obtained by FAM rigid registration, the red dashed line represents the average value of the method, and the blue solid line represents the FAM-TPS FID curve obtained by deformable registration, the blue dashed line indicates the average value of the method.
   \label{figure_7} 
    }  
    \end{center}
\end{figure}
\par
FID and Euclidean distance are used as indicators to evaluate the difference in registration accuracy of FAM and FAM-TPS algorithms for forearms with different rotation angles. FID is a distance index that measures whether two images are similar. From Figure \ref{figure_7}, it can be observed that the average FID distance of the FAM algorithm that introduces thin-plate spline interpolation is slightly larger. This is because the FID distance is used to judge the similarity between the registered fused image and the fixed image in this experiment. The gap between the affine-transformed moving image and the fixed image is smaller, the FID is larger. 
\begin{figure}[hbp]
   \begin{center}
   \includegraphics[width=10cm]{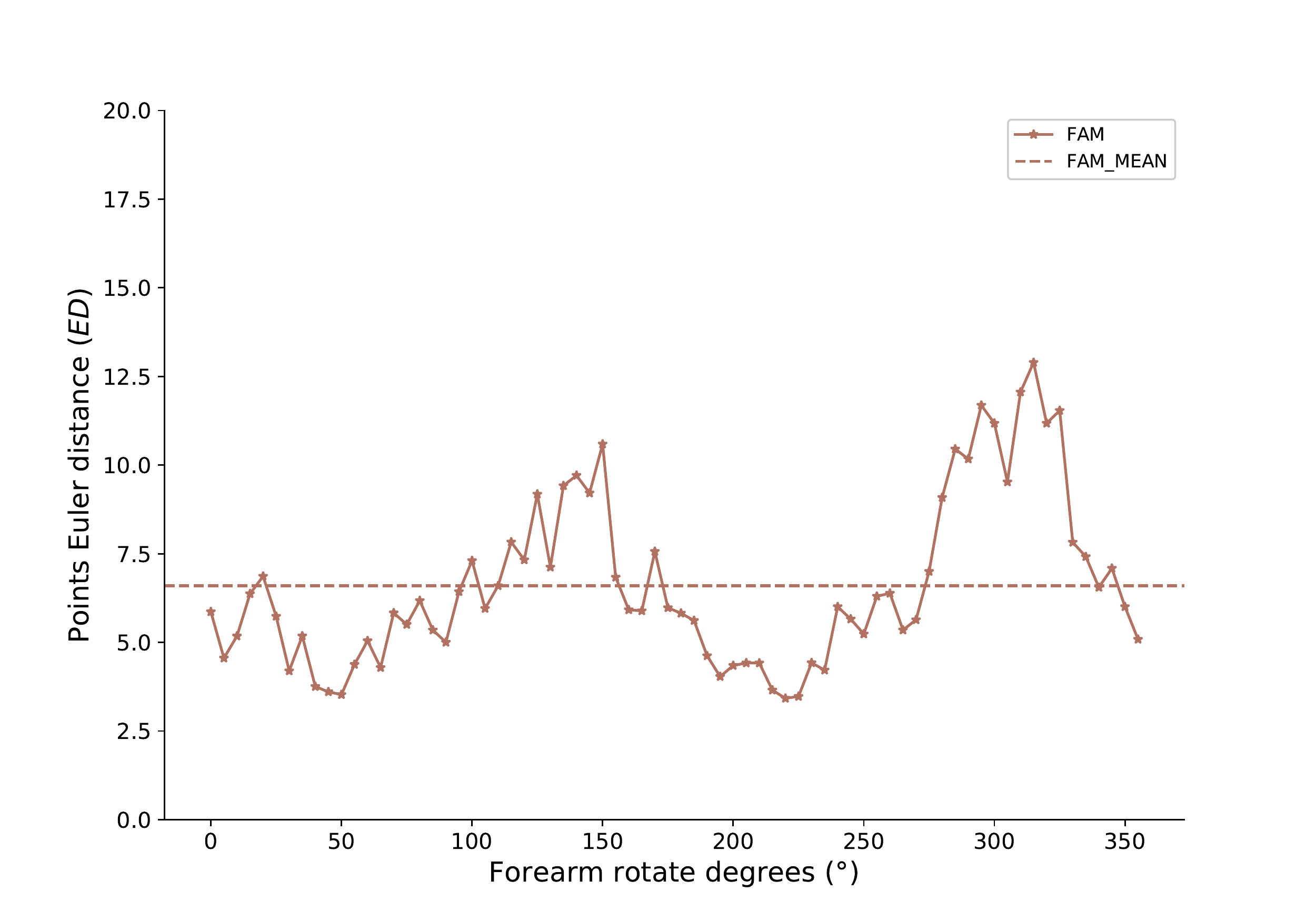}
   \captionv{12}{Short title - can be blank}
   {\textbf{The ED registration evaluation result of the forearm.} The figure shows a curve based on the structure-based feature point extraction accuracy of the forearm. The abscissa represents the axial rotation angle of the forearm, and the ordinate represents the Euler distance between the affine transformation projection point and the real feature point.
   \label{figure_8} 
    }  
    \end{center}
\end{figure}

\par
In addition, we perform the affine transformation on the 20 feature points of the moving extracted by FAM, project them into the fixed image, and calculate the Euclidean distance with the ground-truth to evaluate the matching accuracy of FAM to forearm images. As shown in Figure \ref{figure_8}, it can be observed that when the forearm rotates axially, the projection error of the projection matrix calculated by FAM is very small, and the average value is within 7.5 pixels (the width of the image is 1680 pixels).
\subsection{Relationship between peak value and rotation angle}
\begin{figure}[htbp]
   \begin{center}
   \includegraphics[width=10cm]{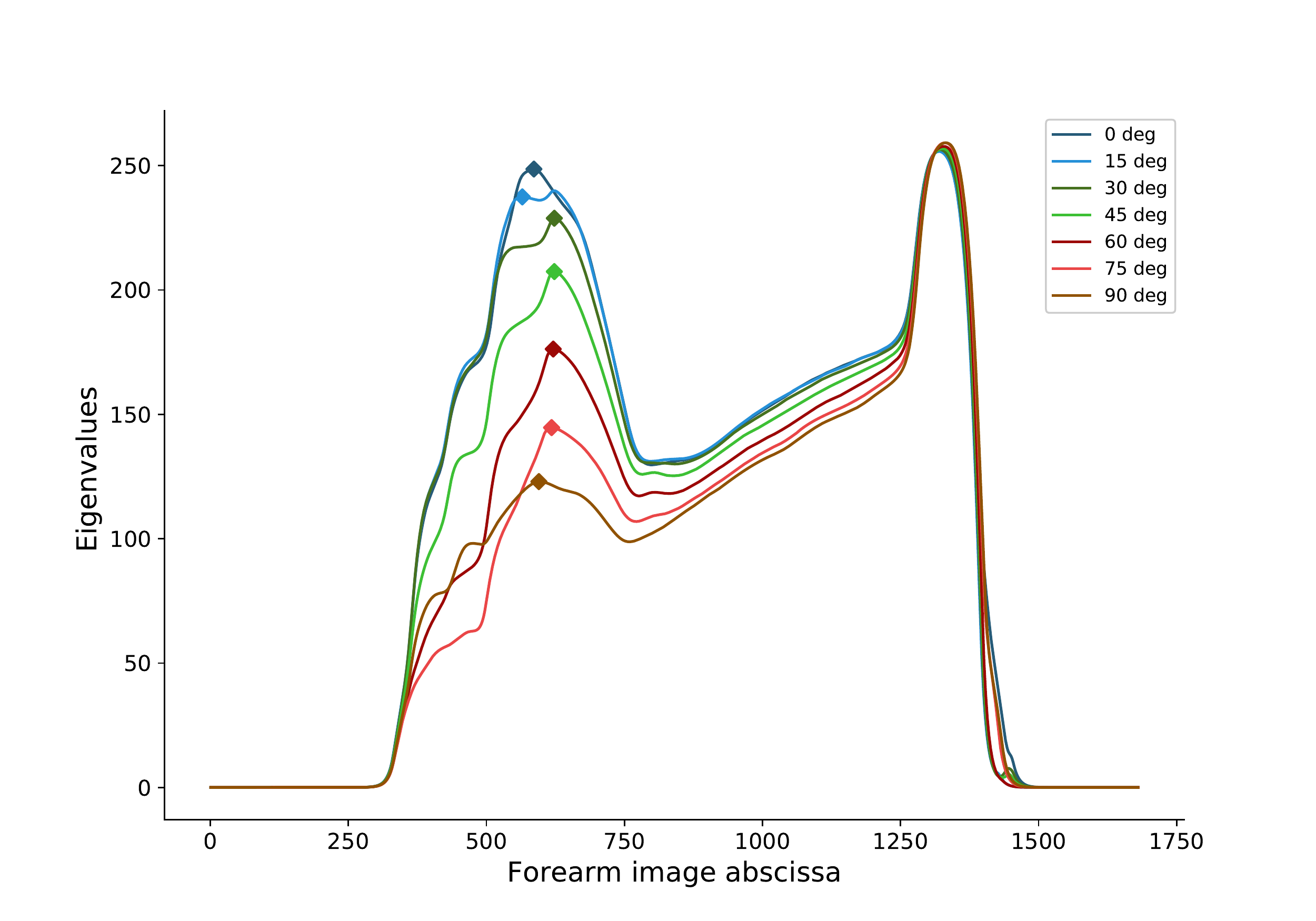}
   \captionv{12}{Short title - can be blank}
   {\textbf{Change of peak of forearm feature curve.} This figure shows the corresponding relationship between the forearm axial rotation angle and the peak value of the feature curve. The curves of different colors correspond to the 0-90° rotation image of the forearm.
   \label{figure_9} 
    }  
    \end{center}
\end{figure}
The relationship between the peak value of the feature curve of the forearm image extracted by FAM and the axial rotation angle of the forearm is shown in Figure \ref{figure_9}. It can be observed that there is a significant correlation between the peak value of the feature curve when the forearm rotates between 0 and 90 degrees. When the forearm rotation angle is larger, the peak value is smaller. Therefore, the approximate rotation angle of the forearm based on the extracted feature curve could be calculated. In addition, by detecting the orientation of the thumb, the axial rotation angle of forearms can be distinguished between 0-90 degrees or 90-180 degrees.


\clearpage 

\section{Discussion}
In this paper, we proposed a structure-based method to extract the key points of the forearm, and provided a framework for multi-modal forearm registration. The framework was tested on a dataset containing 360° axial rotation of the forearm and evaluated using DC, JC, HD, ASD, and ASSD evaluation indicators. In addition, the FID was used to measure the registration similarity. Finally, the relationship between the rotation angle of the forearm and the peak value of the feature curve is given.
\par
\begin{figure}[hbp]
   \begin{center}
   \includegraphics[width=13cm]{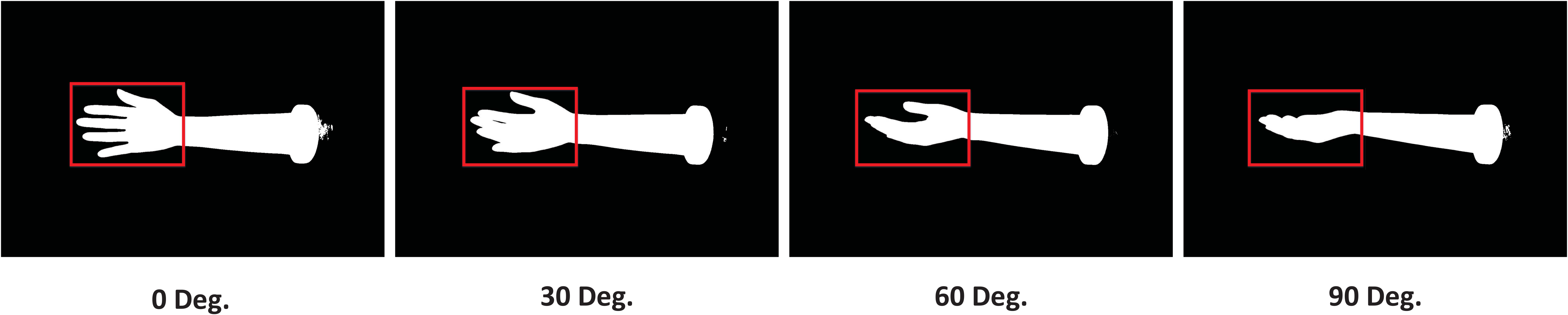}
   \captionv{12}{Short title - can be blank}
   {\textbf{Changes in the projected area of the palm.} This picture shows the change of the forearm binary graph when the forearm axis rotates from 0 degree to 90 degrees. The red box indicates the difference in the change of the palm as it rotates.
   \label{figure_10} 
    }  
    \end{center}
\end{figure}
The structure-based feature point extraction method performs better than other traditional feature extraction-based registration methods on the multi-modal forearm registration dataset. First, we compare the accuracy of the registration results between our algorithm and other classic registration algorithms, as shown in Table \ref{tab_methods_eval}. The results show that the structure-based feature extraction method can reach a higher level in the DC and JC indicators, and can reach a lower level in the HD, ASD, and ASSD distance indicators. Secondly, the registration effect of our algorithm was verified in different images where the forearm was rotated 360 degrees. The FID was used to evaluate the similarity of the registration, and the Euler distance of the feature points was used to evaluate the accuracy of feature point extraction. As shown in Figure \ref{figure_7}, it can be observed that the feature point extraction algorithm based on the structure has better robustness. Finally, when we draw the feature curve of each image, it is observed that the first peak of the feature curve has a clear correlation with the rotation angle of the forearm, as shown in Figure \ref{figure_9}. This is because when the forearm rotates axially, the number of projections in the normal direction of the forearm will change. This change is most obvious in the palm of the hand, as shown in Figure \ref{figure_10}. Therefore, the peak value of the feature curve of the structure-based feature extraction method can correspond to the rotation angle of the forearm.
\par
As shown in Figure \ref{figure_6}, the difficulty of the forearm registration dataset is that the texture of the fixed image is relatively simple, while the texture of the moving image is more complicated. It is difficult to complete multi-modal forearm registration through the registration algorithm based on feature extraction and matching. The structure-based feature point extraction method can quickly locate the wrist position based on the forearm structure, resulting in a matching feature point pair. However, the position of the elbow feature point can not be directly located by the feature curve, and must rely on skin color segmentation to locate the forearm boundary.
\par
We did not consider the corresponding relationship of feature points based on the frames before and after the video in this paper. Therefore, for the registration of the video, we only regard a single frame as an independent image, so the speed of processing the video will be sacrificed. In the future, we will study the feature point position tracking mode based on time information. After the feature point position is determined in the first frame of the video, the image processing of the subsequent frames is omitted, and the feature point extraction is transformed into a tracking problem, which can improve the registration speed. In addition, the non-rigid registration in our algorithm uses thin-plate spline interpolation. Due to the lack of registration and labeling samples, it is difficult for us to train the neural network in our experiment. In the future, we would study some unsupervised learning registration methods based on structural information to further improve registration accuracy.
\par
This research mainly focuses on the registration of multi-modal forearm images. Through the feature point extraction method based on the structure we proposed and thin-plate spline interpolation, a higher level of registration can be achieved. Future work will collect datasets of other limbs and study more robust structure feature extraction methods. At the same time, the algorithm is further accelerated to meet the clinical needs of repairing arm tendons or nerves by C++ implementation.
\clearpage

\section{Conclusion}
We present a structure-based feature point extraction method and use the thin-plate spline interpolation method to achieve multi-modal forearm registration. We verified the accuracy of its registration and its robustness when the forearm rotates in the axial direction, and the relationship between the axial rotation angle of the forearm and the feature peak value is given. With further acceleration, such as C++ implementation, the algorithm has the potential for application in time-sensitive clinical environments, such as forearm tendon or nerve repair surgery.
\clearpage 


\section*{References}
\addcontentsline{toc}{section}{\numberline{}References}
\vspace*{-20mm}










\bibliographystyle{./medphy.bst}    


\end{document}